\documentclass[11pt,a4paper]{article}
\usepackage[hyperref]{acl2021}
\usepackage{times}
\usepackage{latexsym}

\hypersetup{breaklinks=true}

\usepackage{microtype}

\aclfinalcopy

\title{A Survey of Code-switching: Linguistic and Social Perspectives for Language Technologies}

\author{A. Seza Do{\u{g}}ru{\"o}z \\
  Ghent University, Gent, Belgium\\
  \texttt{as.dogruoz@ugent.be} \\
  \And Sunayana Sitaram\\  Microsoft Research India, Bangalore, India\\   \texttt{sunayana.sitaram@microsoft.com}\\
  \AND
  Barbara E. Bullock \\
  UT at Austin, Austin, USA \\
  \texttt{bbullock@austin.utexas.edu}\\  \And Almeida Jacqueline Toribio \\  UT at Austin, Austin, USA \\ \texttt{toribio@austin.utexas.edu} \\}

\date{}

\begin{document}
\maketitle
\begin{abstract}
The analysis of data in which multiple languages are represented has gained popularity among computational linguists in recent years. So far, much of this research focuses mainly on the improvement of computational methods and largely ignores linguistic and social aspects of C-S discussed across a wide range of languages within the long-established literature in linguistics. To fill this gap, we offer a survey of code-switching (C-S) covering the literature in linguistics with a reflection on the key issues in language technologies. 
 From the linguistic perspective, we provide an overview of structural and functional patterns of C-S focusing on the literature from European and Indian contexts as highly multilingual areas. From the language technologies perspective, we discuss how massive language models fail to represent diverse C-S types due to lack of appropriate training data, lack of robust evaluation benchmarks for C-S (across multilingual situations and types of C-S) and lack of end-to-end systems that cover sociolinguistic aspects of C-S as well. Our survey will be a step towards an outcome of mutual benefit for computational scientists and linguists with a shared interest in multilingualism and C-S. 

\end{abstract}

\section{Introduction}
 It is common for individuals in multilingual communities to switch between languages in various ways, in speech and in writing. In example 1, a bilingual child alternates between German and Turkish (in bold) to describe her teacher at school. Note that the Turkish possessive case marker (\emph{-si}) is attached to a German noun \cite{KarakocHerkenrath:19}.
 \begin{enumerate}
    \item  Frau Kummer. Echte Name-\textbf{si} Christa.
    \\Ms. Kummer. Real Name-\textbf{Poss.3sg} Christa.\\
`Ms. Kummer. (Her) real name is Christa'

\end{enumerate}
 The goal of this paper is to inform researchers in computational linguistics (CL) and language technologies about the linguistic and social aspects of code-switching (C-S) found in multilingual contexts (e.g. Europe and India) and how linguists describe and model them. Our intent is to increase clarity and depth in computational investigations of C-S and to bridge the fields so that they might be mutually reinforcing. It is our hope that interested readers can profit from the insights provided by the studies reported in this survey, for instance, in understanding the factors that guide C-S outcomes or in making use of existing annotation schema across multilingual contexts. 

\section{Competing models of C-S}
For linguists, the specific ways in which languages are switched matters. The use of a single Spanish word in an English tweet (ex. 2) is not as syntactically complicated as the integration in ex. 1. In fact, it may not signal multilingualism at all, simply borrowing. Many words, particularly anglicisms, circulate globally: \emph{marketing, feedback, gay}. 
 \begin{enumerate}
 \setcounter{enumi}{1}
    \item This is a good \textbf{baile}!\\
`This is a good dance party!' \cite{solorio2008part}
\end{enumerate}
 To produce example (2), the speaker needs to know only one Spanish word. But, to produce example (1), the speaker has to know what word order and case marker to use, and which languages they should be drawn from. NLP scholars are not always concerned with the difference between examples (1) and (2) so that, with some exceptions \cite{bhat2016grammatical}, grammatical work in NLP tends to rely heavily on the notion of a matrix language model advanced by \citet{joshi1982processing} and later adapted by \citet{myers1997duelling} as the Matrix Language Frame (MLF) model. The MLF holds that one language provides the grammatical frame into which words or phrases from another are embedded and its scope of application is a clause. Thus, it would not apply to the \emph{alternational} English-Afrikaans C-S in example (3) as each clause is in a separate language \cite{van2007grammar}. 
 \begin{enumerate}
 \setcounter{enumi}{2}
     \item  I love Horlicks \textbf{maar hier\'s niks} \\ `I love Horlicks \textbf{but there's nothing there} ' 
 \end{enumerate}
 Although it  dominates computational approaches to C-S, the MLF is contested on empirical and theoretical grounds. The consistent identification of a matrix language is not always possible, the criteria for defining it are ambiguous, and its scope is limited \cite{meakins2012mix, bhat2016grammatical,adamou2016corpus, Macswan:00, auer2005embedded}. \citet{bullock2018predicting} computationally show that different ways of determining the matrix language only reliably converge over sentences with simple insertions as in example (2). 
 
For many linguists, the MLF is not the only way, or even an adequate way, to theorize C-S. The Equivalence Constraint \cite{Poplack:80} captures the fact that C-S tends to occur at points where the linear structures of the contributing languages coincide, as when the languages involved share word order. Other syntactic theories are built on the differences between lexical and functional elements, including  the Government Constraint \cite{disciullo1986code} and the Functional Head Constraint \cite{belazi1994code}. Incorporating the latter in NLP experiments has been shown to improve the accuracy of computational and speech models \cite{li2014language, bhat2016grammatical}. Functional elements include negative particles and  auxiliaries, which are respectively classified as Adverbs and Verbs (lexical classes), in some NLP tag sets \cite{AlGhamdi_2016}. This means that NLP experiments often use annotations that are too coarse to be linguistically informative with regard to C-S. Constraint-free theories \cite{mahootian1996code, Macswan:00} hold that nothing restricts switching apart from the grammatical requirements of the contributing languages. Testing such theories in NLP experiments would require syntactically parsed corpora that are rare for mixed language data \cite{partanen2018dependency}. In sum, working together, theoretical and computational linguists could create better tools for processing C-S than those currently available.

\section{Why do speakers code-switch?}
In addition to focusing on the linguistic aspects and constraints on C-S, linguists are also interested in the social and cognitive motivations for switching across languages. What a (multilingual) speaker is trying to achieve by switching languages can affect its structural outcome. Linguists recognize that pragmatic, interactional, and socio-indexical functions may condition C-S patterns. For instance, \citet{myslin2015code} demonstrate that Czech-English speakers switch to English for high-information content words in prominent prosodic positions when speaking Czech. Other uses of C-S with structural traces include signalling an in-group identity through  \emph{backflagging} \cite{Muysken:1995} or \emph{emblematic} tag-switching \cite{Poplack:80}. These are words or phrases that are used at the edge of clauses (e.g., Spanish \textit{ojal\'a} or English \textit {so}). Other functions, among these, quoting a speaker, getting the attention of an  interlocutor, or reiterating an utterance to soften or intensify a message will also be indicated via C-S in predictable linguistic constructions, such as with verbs of `saying', vocative expressions, and sequential translation equivalents \cite{gumperz1982discourse, alma991018899729706011}. 

According to \citet {Clyne:91}, there are eight factors (e.g. topic, type of interaction, interlocutors, role relationship, communication channel) that can influence C-S choices.  \citet{Lavric:07} explains C-S choices in line with politeness theory, focusing on prestige and face-saving moves in multilingual conversations. \citet{Heller:1992} takes a macro-social view, arguing that French-English C-S in Quebec may signal a political choice among both dominant and subordinate groups.

\citet{Gardner-ChlorosEdwards:04} suggest that social factors influence language choice, with different generations of speakers from the same community exhibiting very different C-S patterns. Similarly \citet{sebba1998congruence} argues that as speakers cognitively construct equivalence between morphemes, words, and phrases across their languages, communities of the same languages may do this differently. Evidence from computational studies suggests that C-S is speaker-dependent \cite{vu2013investigation}. Gender and identity also play a role for C-S practices  in English and Greek Cypriot community in London \cite{Finnis:14}. From a computational perspective, \citet{Papalexakis:14} investigated the factors that influence C-S choices (Turkish-Dutch) in computer mediated interaction and how to predict them automatically.

\section{Code-switching, Borrowing, Transfer, Loan Translation}

While C-S implies active alternation between grammatical systems, borrowing does not. It is difficult to know if a lone word insertion (e.g. example (2)) constitutes a borrowing or a C-S without considering how the items are integrated into the grammar of the receiving language \cite{poplack1988social}. When such analyses are done, most lone-item insertions are analyzable as one-time borrowings, called \emph{nonce borrowings} \cite{sankoff1990}. Similarly, what looks like complex C-S may not be perceived as switching at all. \citet{auer1999codeswitching} distinguishes a continuum of mixing types: prototypical C-S is pragmatic and intentional, \emph{Language Mixing} serves no pragmatic purpose, and \emph{Mixed Languages} are the single code of a community. These can look structurally identical, but the latter can be modeled as a single language  (e.g. languages like Michif Cree \cite{bakker1997language} or Gurinji Kriol \cite{meakins2012mix}) rather than the intertwining of two. \citet{bilaniuk2004typology} describes the Surzhyk spoken by urban Russian-Ukrainian bilinguals (in Ukraine) as `between C-S and Mixed Language' since speakers are highly bilingual and the direction of switching is indeterminate.

Loan translation and transfer involve the words from only one language but the semantics and grammatical constructions from the other. In example 4, the Turkish verb \emph{yapmak},` to do', takes on the Dutch meaning of \emph{doen} in Turkish spoken in the Netherlands \cite{Dogruoz:09}. 

\begin{enumerate}
 \setcounter{enumi}{3}
    \item İlkokul-u İstanbul-da \textbf{yap-tı-m}.  \\
    primary.school-ACC İstanbul-LOC \textbf{do-past-1sg}. \\
`I finished primary school in Istanbul.'
\end{enumerate}

In transfer, grammatical constructions can be borrowed from one language to another without the words being borrowed. \citet{treffers2012grammatical} demonstrates the transfer of verb particles from Germanic languages into French. In Brussels French (Belgium), the construction \emph{chercher après} `look after' (for `look for') is a translation of the Dutch equivalent and, in Ontario French (Canada), \emph{chercher pour}  is the translation equivalent of English `look for'. In reference French (France), there is normally no particle following the verb. The degree to which linguistic features like loan translation and transfer can be found alongside C-S is unknown.

\section{C-S across Languages: European Context}
 The contexts in which people acquire and use multiple languages in Europe are diverse. Some acquire their languages simultaneously from birth, while others acquire them sequentially, either naturally or via explicit instruction. Multilingualism is the norm in many zones where local residents may speak different languages to accommodate their interlocutors. Speakers who use local dialects or minoritized varieties may also be engaged in C-S  when switching between their variety and a dominant one  \cite{mills2015managing, BlomGumperz:1972}. 

C-S in bilingual language acquisition of children has been studied across language contact contexts in Europe. In Germany, \citet{Herkenrath:12} and \citet{Pfaff:99} focused on Turkish-German C-S and \citet{Meisel:94} on German-French C-S of bilingual children. From a comparative perspective, \citet{Poeste:19} analyzed C-S among bilingual, trilingual, and multilingual children growing up in Spain and Germany. In the Netherlands, \citet{BosmaBlom:19} focused on C-S among bilingual Frisian-Dutch children. In addition to analyzing C-S in children's speech, \citet{JuanGarau:01} and \citet{Lanza:98} investigated C-S in the interaction patterns between bilingual children and their parents (i.e. Spanish-Catalan and English-Norwegian respectively). 

Within an educational setting, \citet{Kleeman12} observed C-S among bilingual (North Sami-Norwegian) kindergarten children in the North of Norway. Similarly, \citet{Jorgensen:98} and \citet{Cromdal:04} report the use of C-S for resolving disputes among bilingual (Turkish-Danish) children in Denmark and multilingual (Swedish-English and/or a Non-Scandinavian Language) children in Sweden respectively. 

C-S does not only take place between standard languages but between minority languages and dialects as well. For example, \citet{Themistocleaus:13} studied C-S between Greek and Cypriot Greek and \citet{Deuchar:06} focused on the C-S between Welsh and English in the UK. \citet{Berruto:05} reports cases of language mixing between standard Italian and Italoromance dialects in Italy. In the Balkans, \citet{Kyuchukov:06} analyzed C-S between Turkish-Bulgarian and Romani in Bulgaria. C-S between dialects and/or standard vs. minority languages in computer mediated interaction was analyzed by \citet{Siebenhaar:06} among Swiss-German dialects and by \citet{Robert-Tissot:17} through SMS corpora collected across Germanic (i.e. English and German) and Romance languages (French, Spanish, Italian) in Switzerland.

C-S is commonly observable across immigrant contexts in Europe. In the UK, \citet{Georgakopoulou:09} described the C-S patterns between English and Cypriot Greek while \citet{Issa:06} focused on the C-S between English and Cypriot Turkish communities in London. \citet{Wei:95} analyzed the C-S between English and Chinese from a conversational analysis point of view based on the interactions of bilingual (Chinese-English) families in Northeastern England. In addition, \citet{Ozanska-Ponikwia:16} investigated the Polish-English C-S in the UK as well. C-S among immigrant community members have also been widely studied in Germany (e.g. Turkish-German C-S by \citet{Keim:08} and \citet{cetinoglu-2017-code}, Russian-German C-S  by \citet{Khakimov:16}). In the Netherlands, C-S studies include Turkish-Dutch C-S  by \citet{Backus:10} and Dutch-Morroccan C-S by \citet{Nortier:90}. In Belgium, \citet{Meeuws:98} studied the French-Lingala-Swahili C-S among immigrants of Zaire and \citet{TreffersDaller:94} studied French-Dutch C-S in Brussels. In Spain, \citet{Jieanu:13} describes the Romanian-Spanish C-S among the Romanian immigrants. In addition to the C-S analyses within spoken interactions of immigrant communities across Europe, there are also studies about C-S within computer mediated communication as well. These studies include  Greek-German C-S by \citet{Androutsopoulos:15} in Germany, Turkish-Dutch C-S by \citet{Papalexakis:14}, \citet{Papalexakis:15} and a comparison of Turkish-Dutch and Moroccan-Dutch C-S by \citet{Dorleijn:09} in the Netherlands. Similarly, \citet{Marley:11} compared French-Arabic C-S within computer mediated interaction across Moroccan communities in France and the UK.  

In addition to daily communication, some linguists are also interested in the C-S observed in historical documents. While \citet{Swain:02} explored  Latin-Greek C-S by Cicero (Roman Statesman), \citet{Dunkel:00} analyzed C-S in his communication with Atticus (Roman philosopher who studied in Athens) in the Roman Empire. \citet{Argenter:01} reports cases of language mixing within the Catalan Jewish community (in Spain) in the 14th and 15th centuries and \citet{Rothman:11} highlights the C-S between Italian, Slavic and Turkish in the historical documents about Ottoman-Venetian relations. In Switzerland, \citet{VolkClematide:14} worked on detecting and annotating C-S patterns in diachronic and multilingual (English, French, German, Italian, Romansh and Swiss German) Alpine Heritage corpus. 

Within the media context, \citet{Martin:98} investigated English C-S in written French advertising, and \citet{Onysko:07} investigated the English C-S in German written media through corpus analyses. \citet{Zhiganova:16} indicates that German speakers perceive C-S into English for advertising purposes with both positive and negative consequences. 

Similar to humans, institutions and/or organizations could also have multilingual communication with their members and/or audience. For example, \citet{Wodak:12} analyzed the C-S and language choice at the institutional level for European Union institutions.  

\section{C-S across Languages: Indian Context}

 According to the 2011 Census \cite{chandramouli2011census}, 26\% of the population of India is bilingual, while 7\% is trilingual. There are 121 major languages and 1599 other languages in India, out of which 22 (Assamese, Bangla, Bodo, Dogri, Gujarati, Hindi, Kashmiri, Kannada, Konkani, Maithili, Malayalam, Manipuri, Marathi, Nepali, Oriya, Punjabi, Tamil, Telugu, Sanskrit, Santali, Sindhi, Urdu) are scheduled languages with an official recognition (almost 97\% of the population speaks one of the scheduled languages). Most of the population (~93\%) speak languages from the Indo-Aryan (Hindi, Bengali, Marathi, Urdu, Gujarati, Punjabi, Kashmiri, Rajasthani, Sindhi, Assamese, Maithili, Odia) and Dravidian (Kannada, Malayalam, Telugu, Tamil) language families.  The census excludes languages with a population lower than 10,000 speakers. Given this, it is probably difficult to find monolingual speakers in India considering the linguistic diversity and wide-spread multilingualism. 


\citet{Kachru78} provides one of the early studies on the types and functions of C-S in India with a historical understanding of the multilingual context. In addition to the mutual influences and convergence of Indo-Aryan and Dravidian languages internally, he mentions Persian and English as outside influences on Indian languages. Similarly, \citet{Sridhar78}  provides an excellent comparative overview about the functions of C-S in Kannada in relation to the Perso-Arabic vs. English influences. \citet{Kumar:86} gives examples about the formal (e.g. within NPs, PPs, VPs) and functional (i.e. social and stylistic) aspects of Hindi-English C-S from a theoretical point of view.  More recently, \citet{Doley13} explains how fish mongers in a local fish market in Assam adjust and switch between Assamese, English and local languages strategically to sell their products to multilingual clientele. Another observation about C-S in daily life comes from \citet{Boro20} who provides examples of English, Assamese and Bodo (another language spoken in the Assam region) C-S and borrowings. In addition to English, Portuguese was also in contact with the local languages as a result colonization in South India. For example, \citet{Kapp97} explains the Portuguese influence through borrowings in Dravidian languages (i.e. Kannada and Telugu) spoken in India.

Instead of automatic data collection and methods of analyses, the C-S examples for the above-mentioned studies are (probably) encountered and collected by the authors themselves in daily life interactions over a period of time with limited means. Nowadays, these small sets of data would be regarded as insignificant in computational areas of research. However, ignoring these studies and data could have serious consequences since crucial information about the social and cultural dynamics in a multilingual setting would also be lost. For example, \citet{Nadkarni75} proves this point by explaining how social factors influence the C-S between Saraswat Brahmin dialect of Konkani (Indo-Aryan language) and Kannada (Dravidian language) in the South of India. Both languages have been in contact with each other for over four hundred years. Saraswat Brahmins are fluent in both Konkani and Kannada but they do not speak Konkani with Kannada speakers and they also do not C-S between Konkani and Kannada. \citet{Nadkarni75} attributes this preference to the high prestige associated with Konkani within the given social context. Since Kannada (perceived as less prestigious) is widely spoken in that region, Konkani speakers learn and speak Kannada for functional purposes in daily life which does not involve C-S. However, it is not common for Kannada speakers to learn and speak Konkani \cite{Nadkarni75}.



C-S in India has been investigated through written media, advertising and film industry as well. \citet{Si2011} analyzed Hindi-English C-S in the scripts of seven Bollywood movies which were filmed between 1982 and 2004. Her results indicate a change of direction C-S over the years. More specifically, Hindi was the dominant language with occasional switches to English for the early productions but English became the dominant language especially for younger generations in the later productions. A similar trend has been observed for Bengali movie scripts as well. Through analyzing movie scripts (between 1970s and 2010s), \citet{Chatterjee16} finds a drastic increase in the use of bilingual verbs (e.g. \emph{renovate koreche} ``renovation do") over time and attributes this rise to the increasing popularity of English in Indian society. Within the immigrant context, \citet{Gardner-Chloros:07} focused on the types and functions of C-S between Hindi and English across the TV programs (e.g. highly scripted vs. loosely scripted programs) of a British/Asian cable channel in the UK. Although they have come across C-S in a variety of TV shows, the least amount of C-S was encountered in the news broadcasts (i.e. highly scripted). In general, they have encountered less C-S on TV broadcasts in comparison to the natural speech and attribute this factor to the consciousness of TV personalities about pure language use (instead of C-S). Similarly, \citet{Zipp17} analyzed Gujarati-English C-S within a radio show targeting British South Asians living in the US and concluded that C-S was part of identity construction among youngsters (group identity). \citet{pratapa2017quantitative} perform a quantitative study of 18 recent Bollywood (Hindi) movies and find that C-S is used for establishing identity, social dynamics between characters and the socio-cultural context of the movie.


From an advertising point of view, \citet{kathpalia15} analyzed C-S in Hinglish (i.e. Hindi, English, Urdu, Sanskrit according to their definition) billboards about the Amul brand in India. After compiling the advertisements on billboards (1191), they classified the structures and functions of C-S. Their results indicate more intrasentential C-S than intersentential ones on the billboards. In terms of function, the advertisers used C-S to indicate figures of speech (e.g. puns, associations, contradictory associations, word-creation and repetitions) to attract the attention of the target group.

\citet{Mohanty06} provides an extended overview of the multilingual education system in India exploring the types and quality of schools across a wide spectrum. In general, high-cost English Medium (EM) education is valued by upper-class and affluent families. Although low-cost EM education is also available for lower income families, he questions its impact in comparison to education in the local languages. \citet{Sridhar02} explains that C-S is commonly practiced among students in schools across India. In addition, she finds it unrealistic to ask the students to separate the two languages harshly. In immigrant contexts, \citet{Martin06} investigates how Gujarati-English C-S is used among the South Asian students in educational settings in the UK. Another analysis reveals a shift from Bengali toward English among the younger generations of the immigrant Bengali community in the UK \cite{Alazami06}. In terms of the C-S patterns, first generation immigrants integrate English words while speaking Bengali whereas English dominates the conversations of younger generations with occasional switches to Bengali. There are also studies about Bengali-English C-S in the UK school settings \cite{Pagett06} and Bangladesh \cite{Obaidullah16} as well. However, a systematic comparison between Bengali-English C-S in India, Bangladesh and immigrant settings are lacking. 


In their study about aphasic patients, \citet{Chengappa04} report increased frequency of C-S between Malayalam and English for aphasic patients in comparison to the control group. However, there were less differences between the groups in terms of functions of C-S. \citet{Deepa19} find that amount and types of C-S could be used to differentiate between healthy and mild dementia patients who are bilingual in Kannada and English. Although both studies are carried out with limited subjects, they offer insights about the use of C-S in health settings as well.

\section{Computational Approaches to C-S}

There has been significant interest in building language technologies for code-switched languages over the last few years, spanning a diverse range of tasks such as Language Identification, Part of Speech Tagging, Sentiment Analysis and Automatic Speech Recognition. In the European language context, work has mainly focused on Turkish-Dutch, Frisian-Dutch, Turkish-German and Ukranian-Russian with some initial attempts being made in parsing Russian-Komi text. In the Indian language context, Hindi-English is the most widely studied language pair for computational processing, with some recent work on Telugu-English, Tamil-English, Bengali-English and Gujarati-English. \citet{sitaram2019survey} provide a comprehensive survey of research in computational processing of C-S text and speech and \citet{jose2020survey} present a list of datasets available for C-S research. However, despite significant efforts, language technologies are not yet capable of processing C-S as seamlessly as monolingual data. We identify three main limitations of the current state of computational processing of C-S: data, evaluation and user-facing applications.

\subsection{Data}
The use of Deep Neural Networks, which require large amounts of labeled and unlabeled training data have become the de facto standard for building speech and NLP systems. Since C-S languages tend to be low resourced, building Deep Learning-based models is challenging due to the lack of large C-S datasets. Massive multilingual Language Models (LMs) such as multilingual BERT \cite{devlin2019bert} and XLM-R \cite{conneau2020unsupervised} have shown promise in enabling the coverage of low-resource languages without any labeled data by using the zero-shot framework. These LMs are typically trained in two phases: a ``pre-training" phase, in which unlabeled data from one or multiple languages may be used and a ``fine-tuning" phase, in which task-specific labeled data is used to build a system capable of solving the task.

Since multilingual LMs are trained on multiple languages at the same time, it has been suggested that these models may be capable of processing C-S text \cite{johnson2017google}, with promising results initially reported on POS tagging \cite{pires2019multilingual}. \citet{khanuja2020gluecos} found that multilingual BERT outperforms older task-specific models on C-S tasks, however, the performance on C-S is much worse than the performance on the same tasks in a monolingual setting. Further, these LMs are either trained primarily on monolingual datasets such as Wikipedia in the case of mBERT, or Common Crawl \footnote{http://www.commoncrawl.org} in the case of XLM-R. So, they are either not exposed to C-S data at all during training, or they miss out on several language pairs, types and functions of C-S that are encountered in daily life but not available on the web. 


Since massive multilingual LMs are now replacing traditional models across many NLP applications, it is crucial to consider how they can be trained on C-S data, or made to work for C-S by incorporating other sources of knowledge.

\subsection{Evaluation}

Much of speech and NLP research is now driven by standard benchmarks that evaluate models across multiple tasks and languages. Due to the shortage of standardized datasets for C-S, until recently, the evaluation of C-S models was performed over individual tasks and language pairs. \citet{khanuja2020gluecos} and \citet{aguilar2020lince} proposed the first evaluation benchmarks for C-S that span multiple tasks in multiple language pairs. The GLUECoS benchmark \cite{khanuja2020gluecos} consists of the following C-S tasks in Spanish-English and Hindi-English: Language Identification (LID), Part of Speech (POS) tagging, Named Entity Recognition (NER), Sentiment Analysis, Question Answering and Natural Language Inference (NLI). The LINCE benchmark \cite{aguilar2020lince} covers Language Identification, Named Entity Recognition, Part-of-Speech Tagging, and Sentiment Analysis in four language pairs: Spanish-English, Nepali-English, Hindi-English, and Modern Standard Arabic-Egyptian Arabic. 

Although these benchmarks are important starting points for C-S, it is clear that they do not represent the entire spectrum of C-S, both from the point of view of potential applications and language pairs. Further, it is important to note that while state-of-the-art models perform well on tasks such as LID, POS tagging and NER, they are only slightly better than chance when it comes to harder tasks like NLI, showing that current models are not capable of processing C-S language. Moreover, many of the C-S tasks in the benchmarks above consist of annotated tweets, which only represent a certain type of C-S. Due to these limitations, we currently do not have an accurate picture of how well models are able to handle C-S.

\subsection{User-facing applications}

Although speech and NLP models for C-S have been built for various applications, a major limitation of the work done so far in computational processing of C-S is the lack of end-to-end user-facing applications that interact directly with users in multilingual communities. For example, there is no widely-used spoken dialogue system that can understand as well as produce code-switched speech, although some voice assistants may recognize and produce C-S in limited scenarios in some locales. Although computational implementations of grammatical models of C-S exist \cite{bhat2016grammatical}, they do not necessarily generate natural C-S utterances that a bilingual speaker would produce \cite{pratapa2018language}. Most crucially, current computational approaches to C-S language technologies do not usually take into account the linguistic and social factors that influence why and when speakers/users choose to code-switch.

\citet{bawa2020multilingual} conducted a Wizard-of-Oz study using a Hindi-English chatbot and found that not only did bilingual users prefer chatbots that could code-switch, they also showed a preference towards bots that mimicked their own C-S patterns. \citet{rudra2016understanding} report a study on 430k tweets from Hindi-English bilingual users and find that Hindi is preferred for the expression of negative sentiment. In a follow-up study, \citet{agarwal2017may} find that Hindi is the preferred language for swearing in Hindi-English C-S tweets, and swearing may be a motivating factor for users to switch to Hindi. The study also finds a gender difference, with women preferring to swear in English more often than Hindi. Such studies indicate that multilingual chatbots and intelligent agents need to be able to adapt to users' linguistic styles, while also being capable of determining when and how to code-switch.


Due to the paucity of user-facing systems and standard benchmarks covering only a handful of simpler NLP tasks, it is likely that we overestimate how well computational models are able to handle C-S. In sum, language technologies for C-S seem to be constrained by the lack of availability of diverse C-S training data, evaluation benchmarks and the absence of user-facing applications. They need to go beyond pattern recognition and grammatical constraints of C-S in order to process and produce C-S the way humans do. Hence, it is important for the CL community to be aware of the vast literature around C-S in linguistics, particularly as we proceed to solving more challenging tasks.





\section{Conclusion}
The goal of this paper was to inform computational linguists and language technologists about the linguistic and social aspects C-S studies focusing on the European and Indian multilingual contexts. 
 There are some similarities (e.g. themes for linguistic research in C-S) but also differences between the two contexts in terms of the social, cultural and historical characteristics. For example, C-S in immigrant communities has been a common theme for both multilingual contexts. In Europe, C-S has been widely studied within the immigrant communities who have come through labor immigration in the 1960s. However, there is a need for more research about the C-S in immigrant languages with a more recent history as well as minority languages and regional dialects. Analyzing C-S in the immigration context is even more complex for Indian languages. There are hardly any systematic linguistic comparisons between the C-S within the same language pairs in India and immigrant contexts (e.g. C-S between Hindi-English in India vs. Hindi-English in the US/UK). There is also a need for more research about C-S between less known language pairs in India. However, some of these languages are not even officially listed (e.g. in census results) since they have less than 10,000 speakers. In these cases, collecting and analyzing the multilingual and C-S data becomes more difficult. 

A common flaw that is shared both by linguistics and computational areas of research is to focus only on the positive evidence and assume that C-S will occur in all multilingual contexts. However, there is also a need for negative evidence to falsify this assumption. As illustrated through an example from Konkani-Kannada language contact in India (see section 6), bilingual speakers may prefer not to C-S due to historical, social and cultural factors operating in that setting. Therefore, developing an automatic C-S system for a random pair of languages without an in-depth and systematic analysis of linguistic and social aspects of C-S in a particular context would not be very useful for the targeted users and/or language technologists. 


To date, the literature focusing on the social and linguistic aspects of  C-S is less visible for CL researchers. This lack of visibility leads to misunderstandings and/or incomplete citations of earlier research which would have saved time and resources for CL research if resolved. One of the reasons is perhaps the differences in publishing traditions between humanities and computational areas of research. Conference and workshop proceedings are commonly accepted means of publication in computational linguistics. Whereas, journal publications, books and/or chapters are the publication forms in humanities. However, guidelines about how to cite publications in humanities are often missing in computational venues. There are also differences in terms of length, review cycles and open access policies between the two fields which may influence the visibility of research output for each other. It is perhaps useful to remember that science advances by standing on the shoulders of giants (i.e. building upon earlier research). With our contribution to the conference, we hope to connect the two fields and start a scientific dialogue to enhance the advancement in both fields. 


\bibliography{anthology,acl2020}
\bibliographystyle{acl_natbib}

\appendix

\end{document}